\documentclass[runningheads]{llncs}

\newif\ifarxiv
\arxivtrue

\usepackage{eccv}

\usepackage{eccvabbrv}

\usepackage{graphicx}
\usepackage{booktabs}

\ifarxiv
\else
\usepackage[accsupp]{axessibility}  
\PassOptionsToPackage{breaklinks,colorlinks,citecolor=eccvblue}{hyperref}
\fi

\ifarxiv
  \usepackage[breaklinks,colorlinks,citecolor=eccvblue]{hyperref}
\else
  \usepackage{hyperref}
  \hypersetup{linkcolor=magenta,urlcolor=magenta}
\fi

\usepackage{algorithm}
\usepackage{algorithmic}
\usepackage{newfloat}
\usepackage{listings}
\usepackage{amsmath}
\usepackage{multirow}
\usepackage[utf8]{inputenc}
\usepackage{booktabs}       
\usepackage{amsfonts}       
\usepackage{nicefrac}       
\usepackage{microtype}      
\usepackage{xcolor}         
\usepackage{colortbl}

\def\ourmodel{PASTEL}

\newcommand{\corrauth}{\textsuperscript{\dag}}

\newlength{\figvspace}
\begin{document}

\title{PASTEL: Panoramic Alignment for Monocular 4D Scene Reconstruction}

\ifarxiv
\titlerunning{PASTEL: Panoramic Alignment for 4D Reconstruction [ECCV 2026]}
\else
\titlerunning{PASTEL: Panoramic Alignment for 4D Reconstruction}
\fi

\author{Yuankun Yang\inst{1} \and
Yi Wei\inst{2} \and
Bo Bai\inst{2} \and
Wenyang Zhou\inst{2} \and
Li Zhang\inst{1}\corrauth}

\authorrunning{Y.~Yang et al.}

\institute{$^{1}$School~of~Data~Science, Fudan~University \\
$^{2}$Central Media Technology Institute, Huawei \\
\vspace{0.9em}
\href{https://github.com/LogosRoboticsGroup/PASTEL}{github.com/LogosRoboticsGroup/PASTEL}}

\maketitle
\begingroup
\renewcommand\thefootnote{}
\footnotetext{%
\corrauth\ Corresponding author:
\href{mailto:lizhangfd@fudan.edu.cn}{lizhangfd@fudan.edu.cn}%
}
\endgroup

\begin{abstract}
Reconstructing 4D scenes from casually captured monocular video is vital for applications in virtual reality (VR) and embodied AI. 
Recent advances in 4D reconstruction and novel view synthesis have substantially propelled this capability. 
However, existing reconstruction methods generally cannot recover regions beyond visible camera limits.
Consequently, we introduce a new paradigm that achieves 4D scene synthesis by combining visible-region reconstruction from monocular input with invisible-region generation beyond observable camera boundaries.
A straightforward approach is to leverage video generation models as ``generative priors'' for invisible-region exploration. 
However, their inherent stochasticity and large solution space prevent stable and consistent view synthesis.
As a result, naively incorporating generative content into the reconstruction process often causes artifacts, especially when camera trajectories deviate significantly from the original video. 
To overcome these challenges, we present Panoramic Alignment for Strategic Exploitation of Generative Priors (\textbf{\ourmodel{}}). 
Specifically, \ourmodel{} proposes panoramic scene alignment, a novel representation that reformulates the intractable 3D "invisible region'' exploration into a tractable 2D directional trajectory planning. 
This is achieved by reducing the viewpoint planning from 6-DoF search to a 2D directional search with explicit visibility boundaries.
By operating within this panoramic space, our method strategically identifies camera trajectories that maximize exploration beyond observable boundaries while minimizing viewpoint deviation. 
Experimental results show that \ourmodel{} can not only extrapolate plausible scene content beyond the observable boundaries of input monocular videos, but also substantially boost monocular 4D reconstruction performance. \ourmodel{} outperforms the previous state-of-the-art method by 0.9dB in full-image PSNR on the DyCheck IPhone dataset.
\end{abstract}

\begin{figure}[ht] 
\centering
\includegraphics[width=\columnwidth]{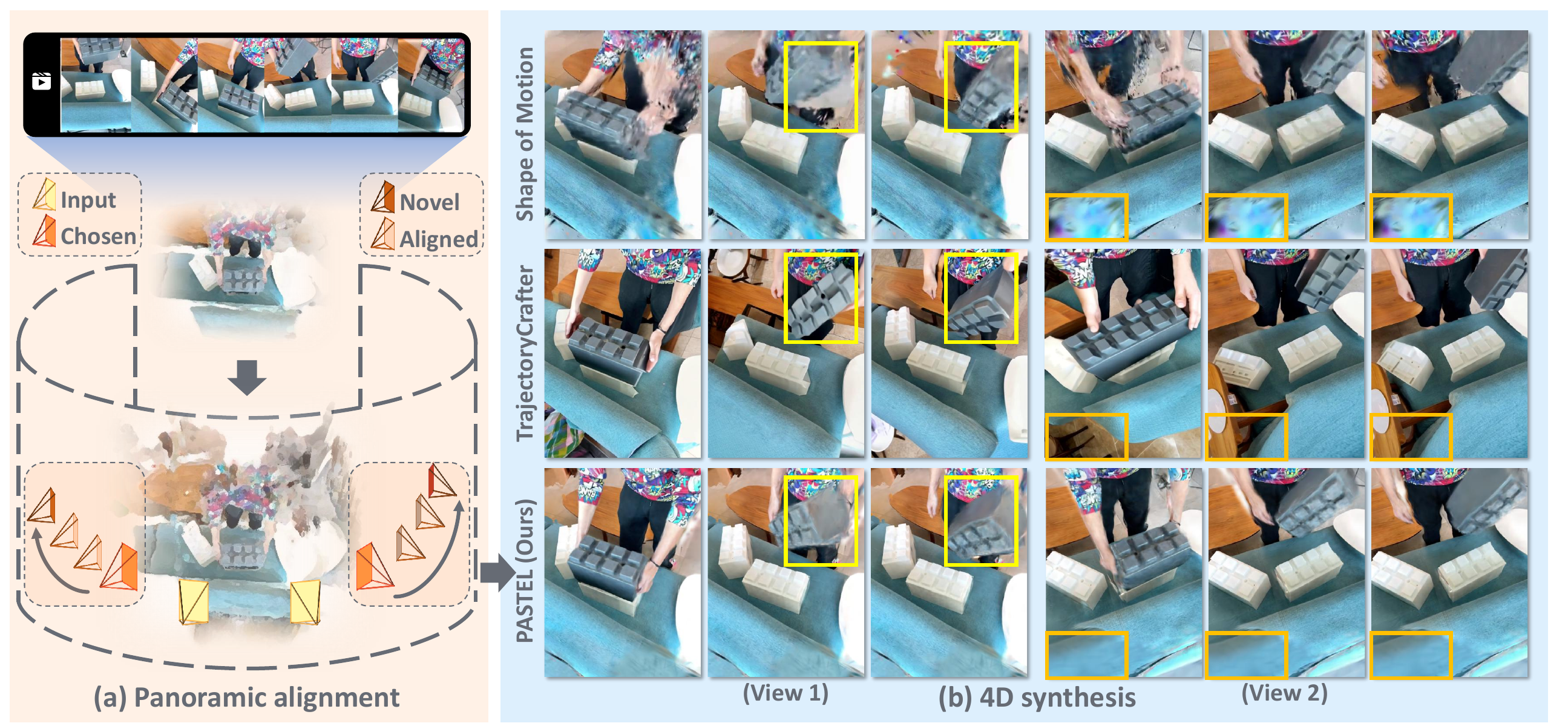} 
\vspace{\figvspace}
\caption{\textbf{Overview of \ourmodel{}}. 
(a) Given monocular videos, \ourmodel{} proposes panoramic alignment to reformulate scene exploration.
(b) \ourmodel{} shows seamless extension of rendering beyond observable boundaries in orange boxed areas. \ourmodel{} also preserves high fidelity and consistency in yellow boxed areas.
}
\label{fig:teaser} 
\vspace{\figvspace}
\end{figure}



\section{Introduction}
Reconstructing geometry-consistent and motion-coherent 4D scenes from monocular video~\cite{li2023dynibar,li2021neural} remains a fundamental and unsolved challenge in computer vision.
Recent advances in novel view synthesis and dynamic scene reconstruction~\cite{park2021nerfies, luiten2024dynamic, wu20244d} have enabled progress in modeling motion dynamics within multi-view settings~\cite{wang2024shape,lei2024mosca}. However, these approaches face substantial limitations when extended to monocular inputs.
This is because monocular videos typically exhibit limited fields of view and severe occlusions. Meanwhile, reconstruction-based methods rely on pixel-level supervision, which cannot infer regions beyond visible camera limits, as shown in the first row of \cref{fig:teaser} (b).
The absence of such ``invisible regions'' introduces a significant gap for novel view synthesis, which limits the immersive quality and user experience of the reconstructed scenes. 

To tackle these ``invisible regions'', an intuitive insight is to effectively leverage camera-controlled generative models~\cite{yu2025trajectorycrafter, you2024nvs,bai2024syncammaster,bai2025recammaster} as ``generative priors'' to infer regions beyond observable camera boundaries.
However, directly applying these methods to reconstruction introduces notable limitations.
Specifically, the inherent randomness within the diffusion denoising process often introduces deviations in the detailed texture and even the overall color. 
This deviation often results in inconsistencies and artifacts when directly applied to the 4D scene reconstruction, as exhibited in the second row in \cref{fig:teaser} (b). 
The greater the trajectory deviates from the original video, the more content needs to be inpainted by the generative prior, resulting in more severe artifacts.
Therefore, designing effective camera trajectories is crucial. Optimal camera trajectories should maximize exploration beyond observable regions while minimizing both trajectory deviation and overlap. 
However, searching for an optimal trajectory of $L$ camera positions entails navigating a parameter space of $L \times 4 \times 4$ entries, since each camera is defined by a $4 \times 4$ extrinsic matrix. This high-dimensional Cartesian search makes the direct optimization of generative trajectories computationally challenging and prone to suboptimal, disjointed viewpoints.

To address this intractable high-dimensional search, we propose Panoramic Alignment for Strategic Exploitation of Generative Priors (\ourmodel). 
Instead of struggling within the unconstrained 3D Cartesian space, our core idea is to reformulate scene exploration by anchoring it within a panoramic domain.
Specifically, we mathematically reformulate the observable 4D scene into a spherical panoramic space. 
This paradigm shift offers a notable advantage: it simplifies the difficulty of high-dimensional camera viewpoint planning. This design reduces the search space from a $L \times 4 \times 4$ Cartesian domain into a 2D directional continuous expansion. 
Within this panoramic space, the highly complex 6-DoF search is substantially simplified. The boundaries of ``invisible regions'' become explicitly quantifiable and globally consistent. 
This elegant formulation enables us to derive expansion trajectories that maximize spatial coverage beyond observable limits while minimizing viewpoint distortion and overlaps.
Subsequently, these naturally continuous panoramic trajectories provide highly stable, geometrically constrained conditioning inputs to any generative prior. 
By aligning the generative inpainting process within this panoramic space, our framework provides geometrically consistent conditioning that effectively constrains the diffusion randomness. This empirically yields comprehensive and geometrically consistent scene expansions.
These generated components then serve as targeted pseudo-supervision to guide the 4D scene reconstruction, successfully mitigating cross-view inconsistencies.

To summarize, we make the following contributions: 
(1) We identify and address a fundamental bottleneck in monocular 4D synthesis: the inherent intractability of view expansions in high-dimensional Cartesian space. To address this, we introduce \ourmodel{}, a conceptually novel framework to promote globally consistent novel view synthesis beyond observable boundaries.
(2) Our core contribution is the panoramic scene alignment, a paradigm shift that reformulates the 4D scene exploration from a $L \times 4 \times 4$ Cartesian search into a constrained, continuous 2D directional field expansion. This formulation elegantly integrates the utilization of generative priors into a continuous space.
(3) Extensive experiments demonstrate that our panoramic alignment seamlessly extends rendering capabilities beyond monocular video coverage while maintaining high fidelity and consistency, establishing new standards in 4D synthesis.
\section{Related works}

\subsubsection{Reconstruction from monocular video}
Video constitutes a primary source of visual data for immersive content creation. Creating immersive experiences and constructing virtual worlds from monocular video holds significant value for applications in VR, robotics, and content creation.
To address this challenge, \cite{lei2024mosca} integrates prior knowledge from multiple 2D foundation models to optimize the proposed motion scaffold representation. 
 \cite{wang2024shape} exploits the low-dimensional structure inherent in scene motion to constrain the full-sequence-long 3D motion of each point.
\cite{li2023dynibar} combines Transformer architectures and image-based rendering to achieve photorealistic novel view synthesis from monocular videos. 
\cite{zhang2024monst3r,han2025d,sucar2025dynamic} employ a data-driven approach, training models to predict the geometry and motions from stereo image pairs.
Some other works extend Gaussian Splatting into dynamic view synthesis~\cite{wu20254d,huang2025vivid4d,li2026gaussian}.
They often incorporate regularization through MLPs~\cite{yang2024deformable} or low-rank motion representations~\cite{huang2024sc}.  
However, these methods rely on pixel-level supervision to reconstruct 3D or 4D scenes. Consequently, they cannot infer ``invisible regions'' where no ground truth is available for supervision.

\subsubsection{Camera-controlled video generation}
Driven by recent advances in video diffusion models, researchers have begun to explore using video generative models for directly synthesizing observations from the altered camera trajectories.
Among them, \cite{you2024nvs,yu2025trajectorycrafter,zhang2024recapture,cao2025uni3c,zheng2026versecrafter} condition the video diffusion models with novel viewpoint cloud rendering.
In contrast, \cite{zhou2025stable,he2024cameractrl,he2025cameractrl} discard the explicit geometry estimation by only conditioning the model with the Plücker embedding of target views.
\cite{bai2024syncammaster,bai2025recammaster} adopt the Diffusion Transformer (DiT)~\cite{peebles2023scalable} architecture with view or frame attention, generating videos from multiple fixed views or a specified trajectory.
Despite the remarkable success in view synthesis, these methods typically exploit denoising process from diffusion models for video generation, which exhibit substantial randomness. This randomness usually leads to severe conflict and inconsistency when directly applied to 3D/4D scene reconstruction.


\subsubsection{Panoramic representations}
Classical approaches assume the panoramic scene as a static image plane.
 \cite{szeliski1997creating, zomet2006seamless} propose cylindrical warping and seam matching to assemble multiple frames onto a unified cylindrical panoramic surface.
 \cite{zaragoza2013projective, zhang2014parallax} introduced parallax-tolerant image stitching processes, relaxing the strict planar assumption but still producing static panoramas. 
To further improve reconstruction of object surfaces, \cite{zheng2007layered} developed layered depth panoramas, and \cite{hedman2017casual, hedman2018instant} introduced mesh-based representations for interactive rendering. Recently, \cite{chugunov2024neural} proposed spherical neural light fields for implicit panoramic image stitching and re-rendering. These advances support vivid wide-angle rendering, but they remain limited by the lack of explicit geometry and physical constraints, making it difficult to represent dynamic object motion and interactions. Unlike panorama stitching for visualization, our work leverages panoramic representations for adaptive trajectory planning and for conditioning off-the-shelf generative priors.

\section{Methods}
\label{sec:method}

\begin{figure*}[t] 
\centering
\includegraphics[width=\textwidth]{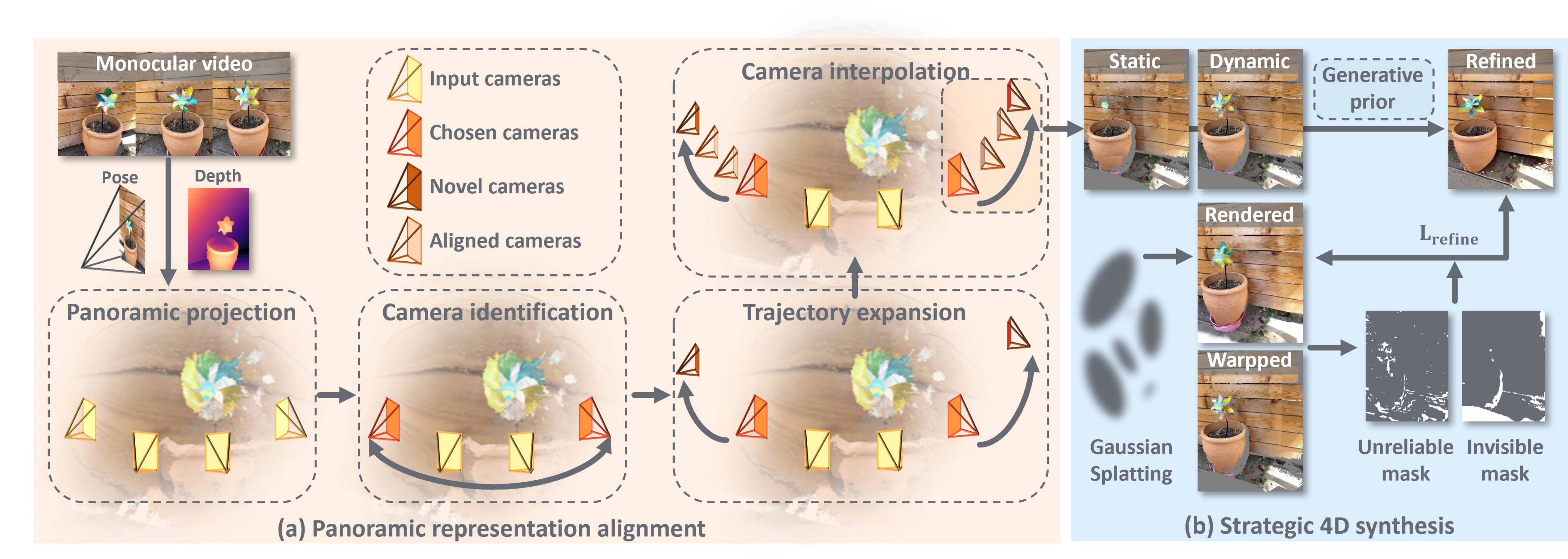} 
\vspace{\figvspace}
\caption{\textbf{Architecture overview of \ourmodel{}}.
(a) Given a monocular video with predicted camera poses and depth, \ourmodel{} designs panoramic representation alignment by projecting all video frames into a unified panoramic space. \ourmodel{} then devises chosen cameras serving as the beginning of trajectories. 
(b) Along the designated trajectory, both static and dynamic components are utilized to achieve a comprehensive view expansion, which feeds into the diffusion prior for refinement. Finally, we compare warped and rendered images from Gaussian Splatting to identify unreliable and invisible regions, which are used to strategically supervise the 4D scene.}
\label{fig:pipeline} 
\vspace{\figvspace}
\end{figure*}

We propose \ourmodel{} for high-quality, unbounded 4D scene synthesis from casually captured monocular videos. 
Instead of relying on a fragmented pipeline of empirical heuristics, \ourmodel{} is built upon a novel panoramic alignment. As illustrated in \cref{fig:pipeline}, the core of our approach is the panoramic reformulation of the 4D scene (\cref{Sec: panorama}). By transforming the Cartesian exploration into a continuous panoramic domain, we can effectively derive expansion trajectories (\cref{Sec: trajectory}). This formulation provides a principled structural constraint for external generative priors, ensuring that the synthesized unobservable regions are intrinsically consistent with the observable scene. 
These geometrically bounded priors then guide the holistic 4D synthesis, yielding high fidelity and cross-view consistency without relying on post-hoc corrections (\cref{Sec: Static-dynamic}).


\subsection{Panoramic scene alignment}
\label{Sec: panorama}
Conventionally, trajectory discovery is performed in full 3D Cartesian space. For $M$ trajectories, each with $L$ viewpoints and full 4×4 extrinsic matrices, the system must optimize $M \times L \times 4 \times 4$ parameters. 
In practice, such empirical optimization often leads to suboptimal scene coverage and visible inconsistencies due to redundant or poorly conditioned viewpoints.
To address this fundamental bottleneck, \ourmodel{} abandons the Cartesian search space entirely. Instead, we introduce a panoramic scene reformulation~\cite{szeliski1997creating, zomet2006seamless} that inherently enforces global geometric consistency and provides an overall view of the scene. 

Given a monocular video of $T$ frames, we first back-project each frame into 3D space using our precomputed depth map, camera extrinsics, and intrinsics. This step produces a sequence of 3D point clouds, each aligned to the world coordinate system. We then compute the spherical center of the panorama $P^{c}=(x^c,y^c,z^c)$. Specifically, to make $P^c$ robust to foreground dynamic objects, we apply depth-percentile filtering to remove the bottom 30\% depth points before computing the mean of the remaining 3D points. We then reproject all 3D points into the panoramic space.
Concretely, for each point in point cloud $p_{t}^{j}=(x_{t}^{j}, y_{t}^{j}, z_{t}^{j}) \in P_{t}$, we first compute the polar coordinates as: 
\begin{align}
   \theta_t^j &= \operatorname{atan2}\left(y_{t}^{j} - y^c, x_{t}^{j} - x^c\right), \\ 
   \quad
   \phi_t^j &= \operatorname{atan2}\left(z_{t}^{j} - z^c, \sqrt{(x_{t}^{j} - x^c)^2 + (y_{t}^{j} - y^c)^2}\right).
\end{align}
Finally, we compute the pixel coordinates within the panorama:
\begin{equation}
   u_t^j = \left(\frac{\theta_t^j}{2\pi} + 0.5\right) \times w_p,
   \quad
   v_t^j = \left(0.5 - \frac{\phi_t^j}{\pi}\right) \times h_p,
\end{equation}
where $w_p$ and $h_p$ are width and height of 2D panorama.
This produces a dense 2D panorama that encodes the global scene. Meanwhile, for each image frame captured by the camera in timestamp $t$, we can also track all its image pixels through the projection. This could yield projected point sets in the panorama for each image, which is denoted as $M_{t} = \{(u_t^j, v_t^j) \, | \, j = 1,..., n_t\}$.

Compared with Cartesian coordinates, the panoramic alignment collapses the high-dimensional trajectory parameterization into a compact spherical coordinate system. This is not merely a coordinate transformation, but a fundamental conceptual shift that makes trajectory optimization mathematically better constrained and explicitly exposes the global topological structure of the scene. 
Within this domain, \ourmodel{} identifies expansion trajectories that enable consistent and efficient novel-view expansion.


\subsection{Adaptive trajectory identification under panorama}
\label{Sec: trajectory}
After projecting the video into panoramic space, \ourmodel{} seeks to determine the optimal camera trajectory under the panoramic space.
Empowered by the panoramic formulation, the 6-DoF camera placement is reparameterized as a guided 2D directional ray-casting. 
This structural simplification enables an interpretable search for the outermost visible boundaries.
The optimal trajectory should balance two goals: minimal deviation from the original cameras and maximal spatial coverage beyond the observed region.
Achieving this balance ensures that generative priors can produce high-quality and consistent inference of regions beyond visible camera limits. 

To ensure uniform exploration with minimum overlap, we determine our trajectory direction $d_m$ through the evenly expanded angle. This is defined as:
\begin{equation}
    d_m = \left( d_{m,u},d_{m,v} \right)= \left( \cos\left(\frac{2\pi \cdot m}{M}\right), \sin\left(\frac{2\pi \cdot m}{M}\right) \right), 
    \label{eq:angle}
    \end{equation}
where $m \in \{1,...,M\}$, $M$ is the total number of evenly expanded directions. Each $d_m$ represents a unit vector in the panoramic plane corresponding to the angle $\frac{2\pi \cdot m}{M}$.
For each direction, we choose a starting camera by finding the point in the panorama that extends furthest along $d_m$. 
In \cref{Sec: panorama}, we have achieved the point sets $M_{t} = \{(u_t^j, v_t^j) \, | \, j = 1,..., n_t\}$ for each image frame captured by the camera in timestamp $t$ under the panorama coordinate. Consequently, we can identify the chosen camera $c_m$ by finding the point with the maximum projection onto direction $d_m$ within each point set $M_t$. 
To prevent the identified trajectory from starting excessively close to or inside a foreground dynamic object which could lead to blocked views or geometrical failures, we introduce a valid search radius constraint $R$ based on the 3D scene depth, which is defined as:
\begin{equation}
R = \mathrm{median}(\|p - P_c\|_2) \cdot \tan(\alpha/2),
\end{equation}
where $\|p - P_c\|_2$ is the Euclidean distance from each 3D point $p$ to the spherical center $P_c$, and $\alpha$ is the angular step of trajectory expansion. This constraint ensures the selected starting point maintains a safe distance from foreground objects. This is done by maximizing the inner product between the relative 2D panorama point sets and the direction $d_m$ within the valid region:
\begin{equation}
   c_m = \arg\max_{t} \left\{ \max_{j \in \Omega_R} \left( \left(u_t^j - \frac{w_p}{2}\right) d_{m,u} + \left(v_t^j - \frac{h_p}{2}\right) d_{m,v} \right)\right\},
    \label{eq:trajectory_interpolation}
\end{equation}
where $\frac{w_p}{2}, \frac{h_p}{2}$ is the center of the 2D panorama. $\Omega_R$ denotes the set of points satisfying the depth constraint $R$.
It ensures that the trajectory begins at the edge of the visible region and naturally expands outward without intersecting foreground objects.

Then, we expand the camera trajectory from the chosen camera $c_m$ through the direction $d_m$ in the panoramic space. The shifted location is then mapped back into Cartesian space, yielding the position of the novel camera. 
The orientation of the novel camera is defined as the direction pointing from the novel camera position toward the spherical center. This design ensures that new viewpoints remain consistent with the scene structure while allowing efficient exploration beyond the observed region.
After that, we create a trajectory by interpolating between the novel camera $(\tilde{R}_{t,m}, \tilde{\mathbf{t}}_{t,m})$ and the original camera $(R_t, \mathbf{t}_t)$. We apply spherical linear interpolation (SLERP) for rotations and linear interpolation for translations:
\begin{equation}
(R_{t,m}^{\theta_l}, \mathbf{t}_{t,m}^{\theta_l}) = \text{Interpolate}\bigl((R_t, \mathbf{t}_t), (\tilde{R}_{t,m}, \tilde{\mathbf{t}}_{t,m}), \theta_l\bigr),
    \label{eq:interpolation}
\end{equation}
where $\{\theta_l\}_{l=1}^{L} \in [0,1]$ denotes the interpolation factor.

By calculating trajectory directions through evenly expanded angles and interpolating between original and new camera positions, we create trajectories with minimized deviation and maximum exploration of areas beyond observable boundaries. This is exclusively enabled by the panoramic representation, which reduces the target trajectory's degrees of freedom from a $L \times 4 \times 4$ space to a directional search.
Crucially, this trajectory design is not a heuristic post-processing step, but a principled structural prior for stabilizing external generative models. It avoids abrupt viewpoint jumps, maintains global scene coherence, and structurally enforces that novel views align smoothly with the physical bounds of the observable scene. As a result, this progression directly suppresses the generative drift inherent to diffusion models.

\subsection{Comprehensive view expansion}
\label{Sec: Static-dynamic}

After deriving the camera trajectory, we leverage it to guide generative priors for 4D synthesis beyond visible camera limits. 
To rigorously bound the stochasticity of the generative prior, we propose a holistic view expansion that naturally integrates static and dynamic elements within the derived panoramic trajectories. 
This alignment preserves coherence with the original video, thereby supporting the generation of more consistent videos.

Specifically, we first reproject the static regions of each frame into 3D point clouds $P_{t}^{\text{static}}$. The static areas are obtained by masking out moving objects using pre-computed dynamic masks from optical flow estimation.
Next, we merge all static point clouds across time to form a denser and wider 3D coverage $\bigcup_{t=1}^{T} P_{t}^{\text{static}}$. This aggregated point cloud is then projected into each target trajectory view, producing static background images $I_{t}^{\text{static}}$.
We then process the moving regions separately. Each frame, together with its depth map, is reprojected onto the target views as $I_{t}^{\text{moving}}$. This provides the dynamic components that align with the designed camera trajectory.
Finally, we combine the static background and the reprojected moving objects as:
\begin{equation}
     I_{t}^{\text{warped}} = I_{t}^{\text{static}} \cup I_{t}^{\text{moving}},
     \label{eq:view_expansion}
\end{equation}
where $t$ denotes different timestamps of the frame.
In this view expansion design, static content from all timestamps can be merged into a unified, viewpoint-consistent background. It prevents the diffusion model from hallucinating inconsistent geometry and provides a stable global reference across the entire 4D sequence.
The resulting warped frames form a trajectory-aligned video $\mathcal{V}^{\text{warped}} = \{I_{t}^{\text{warped}}\}_{t=1}^T$, which is applied as guidance for camera-controlled video re-generation. The generative prior then refines this warped video to produce the refined outputs $\tilde{\mathcal{V}^{\prime}}$.

\ourmodel{} further designs a strategic supervision approach to minimize the adverse effects of RGB estimation deviations from generative priors. The key for strategic supervision is to exploit synthesized videos $\tilde{\mathcal{V}^{\prime}}$ and the previously warped video $\mathcal{V}^{\text{warped}} = \{I_{t}^{\text{warped}}\}_{t=1}^T$ to optimize unreliable and invisible regions of the target 4D scene. 
Specifically, we identify invisible areas as areas not covered by $\{I_{t}^{\text{warped}}\}_{t=1}^T$. Note that we store invalid pixels as -1 during warping, this design could achieve the invisible mask as:
\begin{equation}
    M_{t}^{\text{invisible}} = \mathbf{1}\left( \{I_{t}^{\text{warped}}\}_{t=1}^T < 0 \right).
\end{equation}
We further detect unreliable areas by comparing the Structural Similarity Index (SSIM) between the rendered RGB images $\{I_{t}^{\text{render}}\}_{t=1}^T$ and the warped RGB images $\{I_{t}^{\text{warped}}\}_{t=1}^T$. We set areas where the SSIM falls below a threshold $\epsilon$ as unreliable to achieve the unreliable mask:
\begin{equation}
M_{t}^{\text{unreliable}} = \mathbf{1}\left(\text{SSIM}(I_{t}^{\text{render}}, I_{t}^{\text{warped}}) < \epsilon\right).
\label{eq:unreliable}
\end{equation} 
Finally, we optimize our target scene representation through both the invisible mask and the unreliable mask. The refinement loss is thus defined as:
\begin{equation}
L_{\text{refine}} =  L_{\text{rgb}}( M_{t}^{\text{refine}} \cdot I_{t}^{\text{render}},  M_{t}^{\text{refine}} \cdot  I_{t}^{\text{refined}}), 
\label{eq:refine_loss}
\end{equation}
where $M_{t}^{\text{refine}}=M_{t}^{\text{invisible}} \cup M_{t}^{\text{unreliable}}$ denotes regions for refinement.

By introducing these invisible and unreliable masks, we establish a mutually beneficial collaboration between the reconstruction model and the generative prior. 
The generative prior is strictly confined to inpainting unobserved or severely degraded regions, effectively preventing the diffusion model's stochasticity from overwriting high-fidelity, well-reconstructed areas. Consequently, this strategic supervision effectively mitigates the pervasive artifacts typically caused by unconstrained generative priors in 4D synthesis. Specifically, our pipeline (Fig. 2) depicts step-by-step panoramic projection, trajectory expansion, and supervised distillation with panoramic visualizations. In our implementation, MoSca serves as the underlying 4D representation, while TrajectoryCrafter acts as the generative prior to guide the 4DGS rendering and optimization. We also provide more panoramic visualizations in the supplementary material to clarify the pipeline.


\section{Experiments}

\subsection{Experimental setup}

\label{sec:exp_setup}
\subsubsection{Datasets.}
To quantitatively assess our rendering outcomes both within and beyond the camera's line of sight, we compare our approach to existing methods using the currently most challenging dataset—the DyCheck IPhone~\cite{li2023dynibar}. This dataset provides covisibility masks that encompass regions outside the range of all training cameras, making it ideally suited for a comprehensive evaluation of our performance.  %
We additionally evaluate on Kubric-4D~\cite{greff2022kubric,chen2025reconstruct} to assess view expansion under large camera displacements.
We further conduct experiments on the NSFF dataset~\cite{yoon2020novel} to measure the ability of our method with fewer frames, 
and evaluate the generalization of \ourmodel{} through reconstruction from real-world videos.

\begin{figure*}[t] 
\centering
\includegraphics[width=\textwidth]{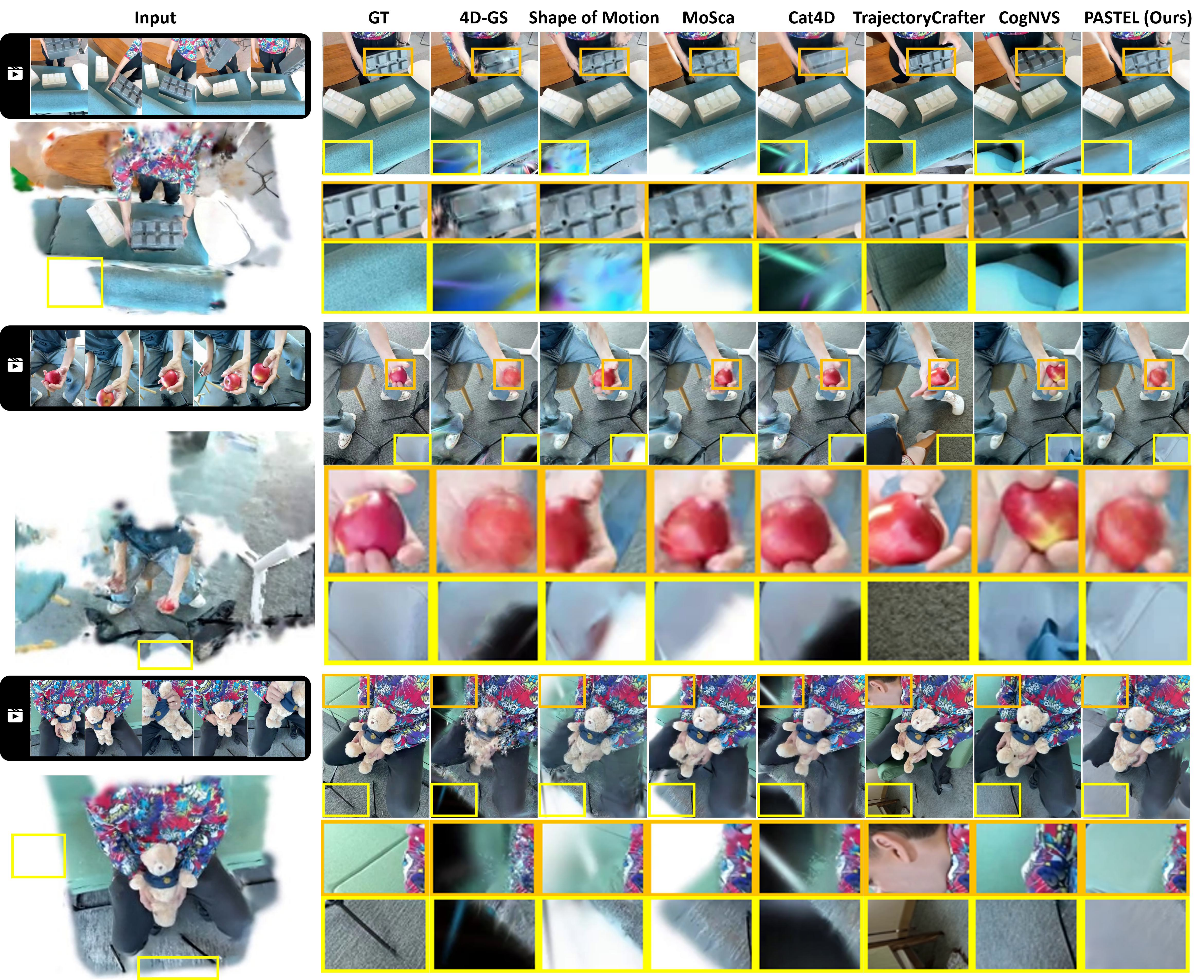} 
\vspace{\figvspace}
\caption{\textbf{Qualitative comparison on DyCheck~\cite{li2023dynibar}.} The yellow boxed areas highlight synthesis beyond covisible regions.
Our method achieves superior overall quality and consistency with the ground truth, excelling in both reconstruction fidelity and the ability to induce details in areas beyond covisible regions.}
\label{fig:results} 
\vspace{\figvspace}
\end{figure*}

\begin{table*}[t]
\centering
\caption{\textbf{Comparison with reconstruction-based and generation-based methods on DyCheck~\cite{li2023dynibar}.} 
``Rec.'' denotes reconstruction time in hours. ``FPS'' denotes rendering speed. 
Our model significantly outperforms all baselines in PSNR, SSIM, and LPIPS, with comparable time cost.}
\label{tab:performance_iphone}
\resizebox{\textwidth}{!}{
\begin{tabular}{l ccc ccc cc}
\toprule
Method & PSNR$\uparrow$ & SSIM$\uparrow$ & LPIPS$\downarrow$ & mPSNR$\uparrow$ & mSSIM$\uparrow$ & mLPIPS$\downarrow$ & Rec. (h) & FPS \\
\midrule
4D GS~\cite{wu20244d}            & 15.71 & 0.450 & 0.398 & 16.54 & 0.594 & 0.347 & 1.2 & 44 \\
Shape-of-Motion~\cite{wang2024shape} & 16.79 & 0.510 & 0.391 & 17.32 & 0.598 & 0.296 & 2.0 & 40 \\
MoSca~\cite{lei2024mosca}        & 17.33 & 0.572 & 0.355 & 19.32 & 0.706 & 0.264 & 1.3 & 38 \\
USPLAT4D~\cite{guo2025uncertainty} & 17.63 & 0.583 & 0.341 & 19.63 & 0.716 & 0.250 & - & - \\
\midrule
Cat4D~\cite{wu2024cat4d}         & 15.91 & 0.427 & 0.398 & 17.39 & 0.607 & 0.341 & - & - \\
TrajectoryCrafter~\cite{yu2025trajectorycrafter} & 12.71 & 0.324 & 0.519 & 14.24 & 0.417 & 0.519 & - & - \\
CogNVS~\cite{chen2025reconstruct} & 15.42 & 0.362 & 0.689 & 16.94 & 0.449 & 0.598 & - & - \\
\midrule
\rowcolor[gray]{.9} 
\ourmodel{} (Ours)       & \textbf{18.54} & \textbf{0.582} & \textbf{0.335} & \textbf{19.75} & \textbf{0.739} & \textbf{0.247} & 1.5 & 38 \\
\bottomrule
\end{tabular}
}
\end{table*}

\subsubsection{Metrics.}
Previous methods evaluated on the DyCheck IPhone dataset~\cite{li2023dynibar} often focus only on areas within covisibility masks, utilizing covisibility masked metrics such as mPSNR, mSSIM, and mLPIPS. However, examining occluded and out-of-view regions is crucial for enhancing user or robotic inference experiences in virtual reality and embodied AI contexts. Therefore, we also evaluate all methods, including areas beyond covisibility masks to assess their ability to extend views outside the camera boundaries and ensure the completeness of 4D scenes, which are denoted as full-image PSNR, SSIM, and LPIPS. Note that standard benchmarks do not provide pixel-level ground truth for purely invisible regions. We thus report metrics as in prior work~\cite{lei2024mosca,guo2025uncertainty}. 

\subsubsection{Implementation details.}
\label{sec:implementation details}
Our panoramic alignment is parameterized in a spherical coordinate system. The width and height of 2D panorama are empirically set as $w_p=2048$ and $h_p=2048$. Each pixel corresponds to a 3D direction vector defined by the azimuthal angle $\theta \in [0, 2\pi)$ and polar angle $\phi \in [-\frac{\pi}{2}, \frac{\pi}{2}]$. This mapping adheres to the standard spherical-to-Cartesian conversion.
Our pipeline builds upon standard estimators: UniDepth~\cite{unidepth} for depth, BootsTAPIR~\cite{doersch2024bootstap} for pose, and RAFT~\cite{raft} for dynamic masks when not precomputed. Nevertheless, our core contribution lies in panoramic alignment rather than these modules. Our generative prior is TrajectoryCrafter~\cite{yu2025trajectorycrafter} (50 inference steps). The 4D scene is represented via Motion Scaffolds~\cite{lei2024mosca}, optimized for 6000 steps. 
A detailed robustness analysis of depth, pose, and generative prior variations is provided in the supplementary material.
We set $\epsilon = 0.3$ for the unreliable mask. The frame length $L=49$ follows~\cite{yu2025trajectorycrafter}. The interpolation factors $\theta_l$ are evenly distributed in $[0,1]$. We set $M=8$ expanded directions to cover extended regions with minimal overlap. Sensitivity to $M$, $L$, and $\epsilon$ is analyzed in the supplementary. Performance remains stable within a broad range around these defaults.
The preprocessing (including depth extraction, dense optical flow, and pose estimation) takes about 5min. The trajectory expansion and TrajectoryCrafter generation take approximately 10min. The final 4D GS reconstruction requires about 1.5h, and the rendering speed reaches 38FPS.

\subsection{Comparison with state of the art}
We compare the performance with both reconstruction-based methods and generative-based methods on the DyCheck IPhone dataset~\cite{li2023dynibar}, the NSFF dataset~\cite{yoon2020novel}, the Kubric-4D dataset~\cite{greff2022kubric}, and other real-world monocular videos.

On the DyCheck IPhone dataset~\cite{li2023dynibar}, the quantitative results are reported in \cref{tab:performance_iphone}, and
qualitative results are shown in \cref{fig:results}.
Reconstruction-based techniques typically excel in regions within the camera's field of view, showing competitive performance in covisibility masked metrics, such as mPSNR, mSSIM, and mLPIPS. However, these methods often lack the capability to interpret occluded regions or areas that fall outside the coverage of covisibility masks, with no prediction of areas beyond the covisibility masks, as illustrated in the red box parts in \cref{fig:results}.
On the other hand, generation-based methods possess the ability to induce regions beyond the covisibility masks but may suffer from inconsistencies in illumination and color fidelity. 
This is attributed to the denoising processes inherent in generative models, which can introduce dataset-specific color biases that distort the original hues.
In comparison, \ourmodel{} demonstrates consistency with monocular video inputs while effectively extending its rendering capabilities to regions that fall outside the coverage of monocular videos in \cref{fig:results}. \ourmodel{}  thus achieves state-of-the-art performance across areas both within and beyond the covisibility mask in \cref{tab:performance_iphone}, outperforming the previous best reconstruction method~\cite{guo2025uncertainty} by 0.9dB in PSNR. 
Moreover, \ourmodel{} achieves the best overall quality on the NSFF dataset~\cite{yoon2020novel}, see supplementary for details.

\begin{figure*}[t]
\centering
\includegraphics[width=\textwidth]{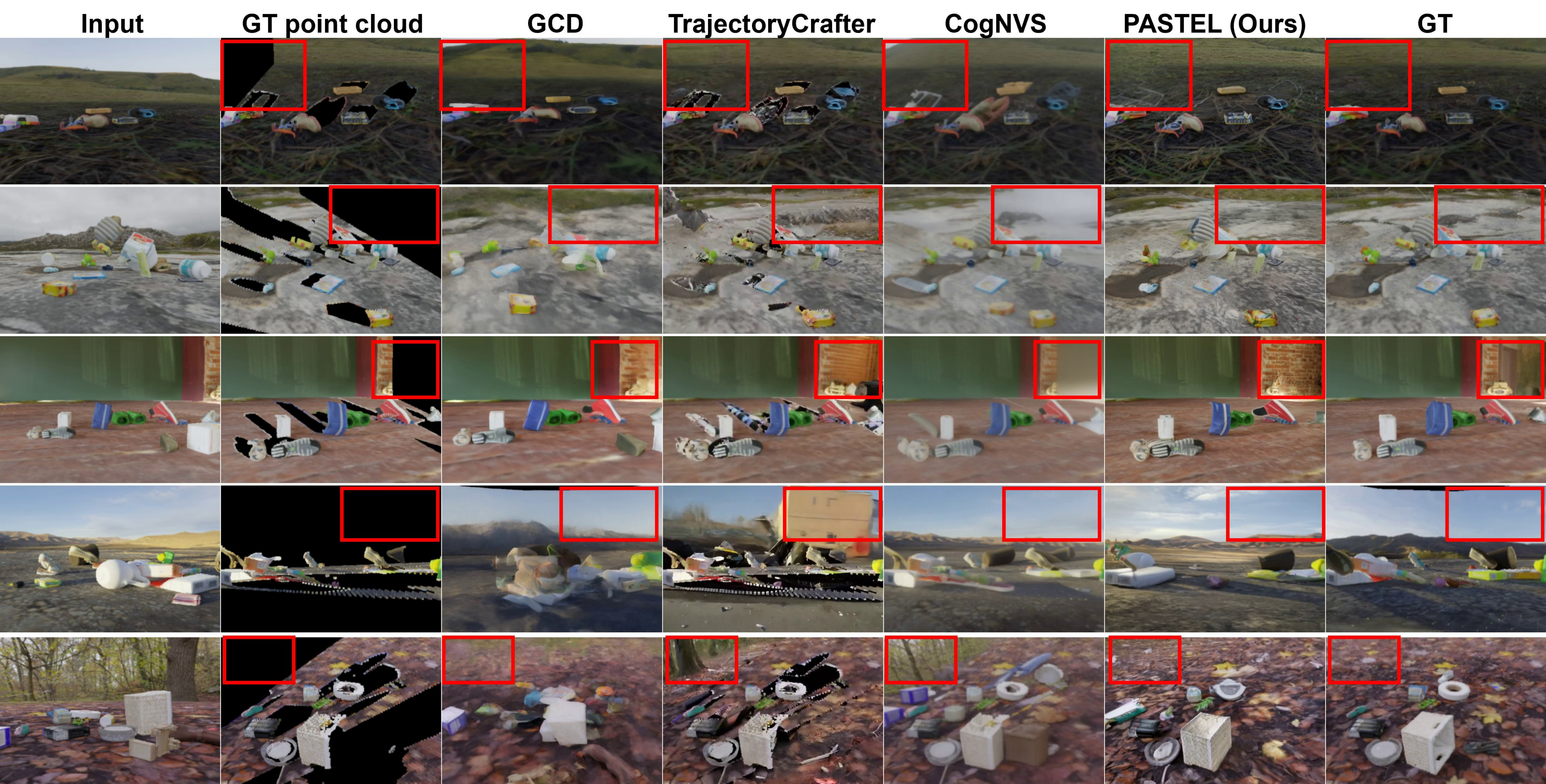}
\vspace{\figvspace}
\caption{\textbf{Qualitative comparison on Kubric-4D~\cite{greff2022kubric}.} The red boxed areas highlight synthesis in previously unobserved regions under large camera displacements.
Our method achieves superior overall quality and consistency with the ground truth.}
\label{fig:results_kubric4d}
\vspace{\figvspace}
\end{figure*}

\begin{table*}[t]
\centering
\caption{\textbf{Comparison with generation-based methods on Kubric-4D~\cite{greff2022kubric}.}
Our model achieves the best full-image PSNR, LPIPS under large camera displacements, while remaining competitive on the remaining metrics.}
\label{tab:kubric4d}
\resizebox{\textwidth}{!}{
\begin{tabular}{l ccc ccc}
\toprule
Method & PSNR$\uparrow$ & SSIM$\uparrow$ & LPIPS$\downarrow$ & mPSNR$\uparrow$ & mSSIM$\uparrow$ & mLPIPS$\downarrow$ \\
\midrule
GCD~\cite{vanhoorick2024gcd} & 18.59 & 0.555 & 0.383 & 18.68 & 0.560 & 0.312 \\
TrajectoryCrafter~\cite{yu2025trajectorycrafter} & 20.93 & 0.730 & 0.257 & 23.44 & 0.781 & 0.186 \\
CogNVS~\cite{chen2025reconstruct} & 22.63 & \textbf{0.760} & 0.232 & \textbf{24.85} & \textbf{0.799} & 0.162 \\
\midrule
\rowcolor[gray]{.9}
\ourmodel{} (Ours) & \textbf{22.75} & 0.733 & \textbf{0.212} & 24.25 & 0.781 & \textbf{0.149} \\
\bottomrule
\end{tabular}
}
\end{table*}

\begin{figure*}[t] 
\centering
\includegraphics[width=\textwidth]{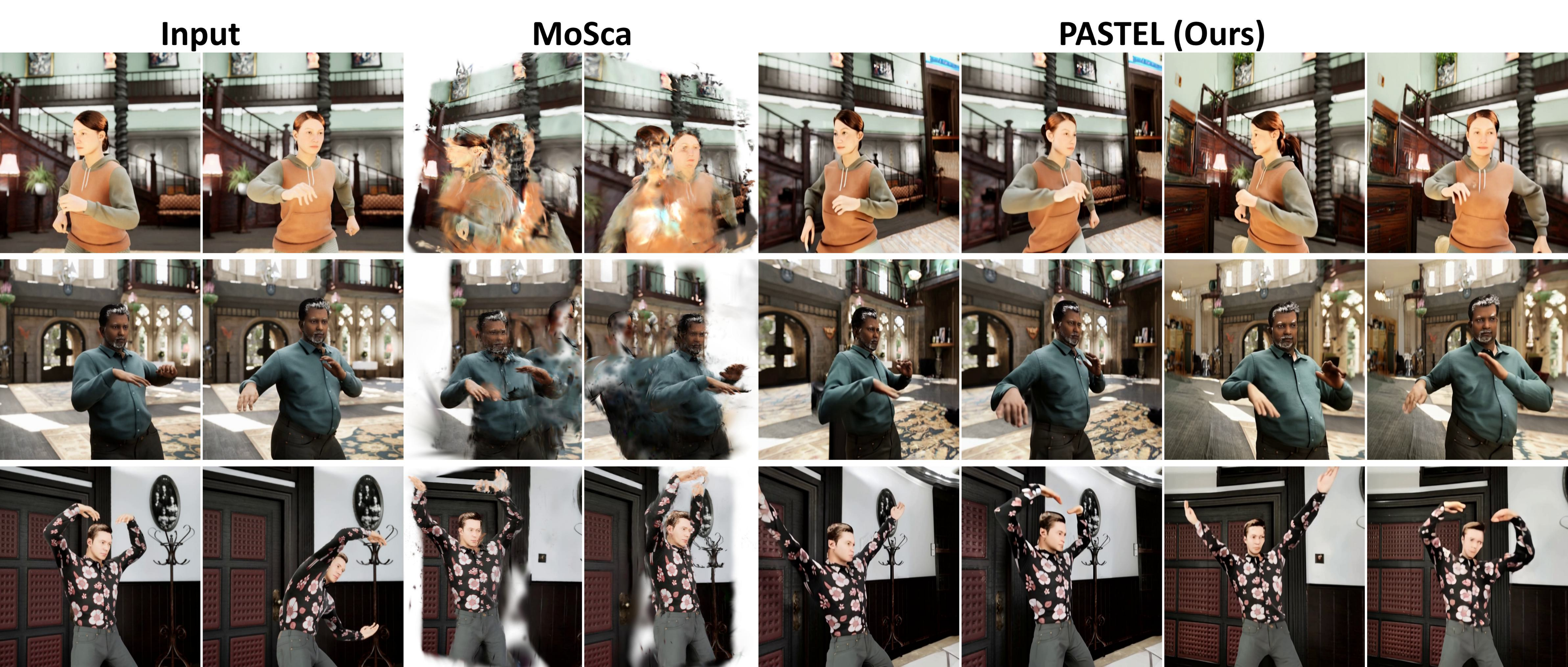} 
\vspace{\figvspace}
\caption{\textbf{ Reconstruction on monocular videos with complex motion and heavy occlusions.}  \ourmodel{} captures highly dynamic scenes with irregular motion, self-occlusion, and dis-occlusion. Compared with MoSca~\cite{lei2024mosca}, \ourmodel{} complements geometry of dynamic content in unseen and occluded regions while improving static backgrounds at the same time.}
\label{fig:extreme} 
\vspace{\figvspace}
\end{figure*}

On Kubric-4D~\cite{greff2022kubric}, the quantitative results are reported in \cref{tab:kubric4d}, and qualitative comparisons are shown in \cref{fig:results_kubric4d}.
This benchmark involves substantially larger camera displacements than DyCheck, providing a more challenging test of view expansion under extreme novel trajectories.
Compared with prior methods in \cref{fig:results_kubric4d}, existing approaches often fail to faithfully complete the red boxed invisible regions under large camera motion, leaving missing geometry, blur, or temporal inconsistency.
Generation-based methods, including GCD~\cite{vanhoorick2024gcd}, TrajectoryCrafter~\cite{yu2025trajectorycrafter}, and CogNVS~\cite{chen2025reconstruct}, can synthesize previously unobserved content, but still struggle to balance perceptual quality with cross-view consistency across the full image.
In comparison, \ourmodel{} combines the cross-view consistency of 4D reconstruction with the inpainting capability of generative priors, enabling coherent synthesis in the red boxed invisible regions while preserving fidelity in observed areas.
As shown in \cref{tab:kubric4d}, \ourmodel{} achieves the best full-image PSNR, while remaining competitive with CogNVS on the remaining metrics.
This demonstrates that our panoramic trajectory planning generalizes effectively beyond the small-baseline setting of DyCheck and supports complete 4D scene synthesis under large camera displacements.
We further evaluate \ourmodel{} on  highly dynamic scenes with significant occlusions and dis-occlusions. As shown in \cref{fig:extreme}, previous state-of-the-art method~\cite{lei2024mosca} often produce missing or broken geometry in the occluded moving body parts. In contrast, \ourmodel{} maintains coherent motion trajectories and complements geometry under irregular human motion and heavy occlusions.
\ourmodel{} also delivers high-fidelity synthesis on other real-world monocular videos with crowded scenes, heavy occlusion, and complex motion. Please refer to the supplementary material for details.

\begin{table*}[t]
\centering
\caption{\textbf{Ablation study on different components of the system.}
Panoramic representation, trajectory alignment, and view expansion are crucial for views beyond the camera range. Strategic supervision markedly improves performance.}
\label{tab:abl_comp}
\resizebox{\textwidth}{!}{
\begin{tabular}{l ccc ccc}
\toprule
Components & PSNR$\uparrow$ & SSIM$\uparrow$ & LPIPS$\downarrow$ & mPSNR$\uparrow$ & mSSIM$\uparrow$ & mLPIPS$\downarrow$ \\
\midrule
\rowcolor[gray]{.9} 
Full model                   & \textbf{18.54} & \textbf{0.582} & \textbf{0.335} & \textbf{19.75} & \textbf{0.739} & \textbf{0.247} \\
w/o panorama (linear)        & 15.73 & 0.497 & 0.370 & 19.42 & 0.728 & 0.249 \\
Cartesian-TopK               & 17.31 & 0.548 & 0.352 & 19.55 & 0.733 & 0.251 \\
w/o trajectory alignment     & 17.87 & 0.572 & 0.342 & 19.57 & 0.736 & 0.247 \\
w/o view expansion           & 17.44 & 0.564 & 0.357 & 19.45 & 0.733 & 0.250 \\
w/o strategic supervision    & 14.09 & 0.306 & 0.548 & 14.80 & 0.506 & 0.445 \\
\bottomrule
\end{tabular}
}
\end{table*}

\subsection{Ablation study}
To demonstrate that our performance gains fundamentally stem from our core conceptual contribution—the panoramic reformulation—rather than merely relying on external generative priors or depth/pose estimators, we systematically ablate the key components of \ourmodel{} on the DyCheck dataset~\cite{li2023dynibar} in \cref{tab:abl_comp} and \cref{fig:ablation}. We further include more ablations on hyper-parameters in the supplementary.
\subsubsection{Panorama.}
\begin{figure*}[t] 
\centering
\includegraphics[width=\textwidth]{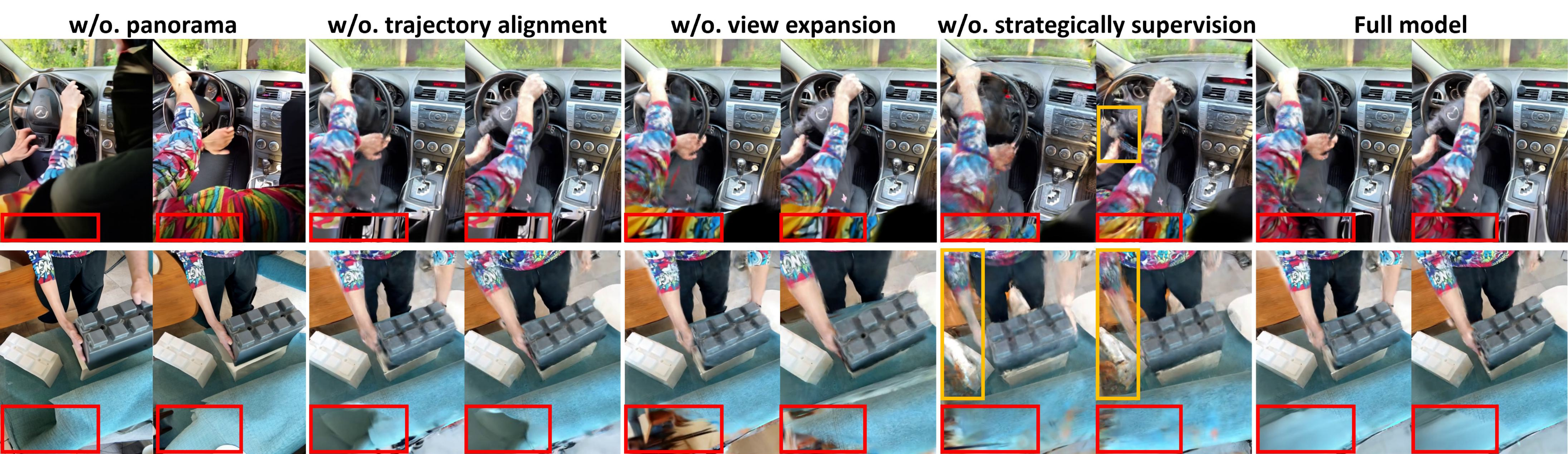} 
\vspace{\figvspace}
\caption{\textbf{Visual comparison of ablations.} As exhibited in the red-boxed areas, panoramic representation facilitates the effective extension of views beyond camera boundaries. Trajectory alignment and view expansion contribute to the consistent synthesis of extended views of red-boxed areas.
Strategic supervision is pivotal in harmonizing the generative content and the original monocular video, culminating in coherent 4D renderings in the orange-boxed areas.}
\vspace{\figvspace}
\label{fig:ablation} 
\end{figure*}

Our core contribution is reformulating the 4D scene reconstruction into a panoramic space. This formulation simplifies the Cartesian camera search into a 2D directional expansion. 
To isolate the contribution of our panoramic formulation, we compare against a linear Cartesian baseline (``w/o panorama (linear)'' in \cref{tab:abl_comp}) and a stronger Cartesian baseline (``Cartesian-TopK'') that selects coverage-maximizing TopK views from 100 random trajectories. Without panoramic projection, trajectory planning must operate in full 3D Cartesian space. We extend trajectories linearly from the first frame for the linear baseline, as no existing method provides structured directional planning for this task.
As shown in \cref{fig:ablation} and \cref{tab:abl_comp}, without our panoramic reformulation, the identical generative prior collapses under the unconstrained Cartesian search space, leading to suboptimal coverage and a drop in all quantitative metrics. Even with the stronger Cartesian-TopK sampling strategy, PASTEL still outperforms it significantly, confirming the panoramic formulation's advantage beyond sampling strategy. The panorama provides structured trajectory planning and geometric warping priors, not just camera placement.
The substantial performance gap between our full model and these Cartesian baselines strongly supports that the strong performance of \ourmodel{} is unlocked by our panoramic framework. 

\subsubsection{Trajectory alignment.}
Rather than being an isolated engineering step, trajectory alignment is a natural extension derived from our panoramic formulation. It ensures that the newly synthesized viewpoints evolve continuously within the spherical space. This continuous alignment intrinsically bounds the generative prior by avoiding abrupt viewpoint jumps, which improves PSNR, SSIM, and LPIPS across the entire image independently of the prior's specific architecture.

\subsubsection{View expansion.}
The view-expansion mechanism in \cref{eq:view_expansion} leverages the globally consistent nature of our panoramic domain to merge static content from all timestamps into a unified background. This cohesive representation naturally ensures consistent inference for extended views, yielding substantial improvements in all metrics and further demonstrating the intrinsic value of operating in the panoramic space.

\subsubsection{Strategic supervision.}
Our framework does not indiscriminately absorb the outputs of external generative modules. Instead, the strategic-supervision design in \cref{eq:refine_loss} acts as a rigorous filter, selectively applying generative priors only to invisible and unreliable regions based on geometric confidence. This principled approach is pivotal in reconciling the inherent hallucinations of diffusion models with the physics of the original monocular video. As demonstrated in \cref{fig:ablation} and \cref{tab:abl_comp}, this strategy effectively quarantines generative errors, explicitly proving that our carefully designed framework is responsible for maintaining the physical integrity and geometric coherence of the final 4D rendering.

\section{Conclusion}
In this paper, we introduce \ourmodel{}, a framework for monocular 4D scene synthesis that combines visible-region reconstruction with invisible-region generation beyond observable camera boundaries. Our approach employs Panoramic Alignment for Strategic Exploitation of Generative Priors, overcoming the cross-view inconsistency and stochasticity typical of generative models while extending rendering beyond monocular coverage. Central to our approach is the panoramic representation, which simplifies unconstrained 3D Cartesian scene exploration into a structured 2D search. This enables the effective planning of target cameras and trajectories. It optimizes the conditioning for generative priors and ensures high-fidelity inference beyond observable boundaries.
We anticipate that the unbounded 4D scene synthesis achieved by \ourmodel{} will unlock significant potential for immersive Virtual Reality experiences and  embodied AI.

\section*{\ackname}
This work was supported in part by New Generation Artificial Intelligence-National Science and Technology Major Project (2025ZD0123004), Ningbo grant (2025Z038) and National Natural Science Foundation of China (Grant \\ No.~62376060).

%
%
\bibliographystyle{splncs04}
\bibliography{main}

\ifarxiv
\clearpage
\appendix

\section{Preliminaries on Camera-Controlled Video Re-generation}
\label{Sec: video re-generation}

Camera-controlled video re-generation aims to redirect the camera trajectory of input monocular video for creating immersive exploration.
Because of the diffusion model's desirable diversity and scalability, it has become the \textit{de facto} solution for various visual generation tasks.
Given a video $\mathcal{V}_0 \in \mathbb{R}^{T \times C \times H \times W}$, the forward diffusion process gradually adds noise to it as:
\begin{equation}
    q(\mathcal{V}_{t}|\mathcal{V}_{t-1}) = \mathcal{N} (\mathcal{V}_{t}; \sqrt{1-\beta_{t}} \mathcal{V}_{t-1}, \beta_t \mathbf{I}),
\end{equation}
where $\beta_{t} \in (0,1)$ represents the pre-defined variance level at time step $t$, $\mathcal{V}_{1}$ will degrade into pure Gaussian noise.
The diffusion model is proposed to learn the reverse process, thus being able to sample from the data distribution by denoising pure Gaussian noise.
The reverse process can be formulated as:
\begin{equation}
    \mathcal{V}_{t-1} = \frac{1}{\sqrt{\alpha_{t}}} \left(
    \mathcal{V}_{t} - \frac{\beta_{t}}{\sqrt{1-\bar{\alpha}_t}}\epsilon_{\theta} \left( \mathcal{V}_{t}, t \right)
    \right),
\end{equation}
where $\bar{\alpha}_t = \prod_{s=1}^{t} (1-\beta_s)$, $\epsilon_{\theta}$ is a noise estimator, modeled as a Diffusion Transformer (DiT). To achieve affordable computation for long and high-resolution video, the diffusion process is conducted in the latent space, which can be transformed between pixel space via a 3D VAE.
Since our task is conditioned on the original trajectory video $\mathcal{V}^{\prime}$, $\epsilon_{\theta}$ can also take $\mathcal{V}^{\prime}$ as input by concatenating its warped version with the noisy latent and directly referring to it through cross attentions.

\subsection{Challenges in Cartesian Optimization}
Despite impressive perceptual quality, video diffusion alone is not sufficient for real-world applications that require real-time rendering and consistent outputs. This limitation motivates us to explore distilling video generation priors into a structured 4D scene representation.
However, the inherent randomness of the diffusion denoising process often introduces deviations in both detailed textures and global colors. These deviations lead to inconsistencies and artifacts when directly used for 4D scene reconstruction, which become extremely severe when the synthesized camera trajectories deviate strongly from the original video.
Consequently, designing effective camera trajectories is crucial. The trajectories should expand exploration beyond the observable regions while minimizing both deviation and overlap. However, trajectory optimization in the Cartesian coordinate is difficult: a trajectory with $L$ camera positions involves $L \times 4 \times 4$ degrees of freedom. This high dimensionality makes direct search for optimal trajectories intractable in the Cartesian coordinate.

\begin{table}[t]
\centering
\caption{\textbf{Quantitative comparison with reconstruction-based methods on Nvidia~\cite{li2023dynibar}.}
The best results are highlighted in \textbf{bold}, while the second best are \underline{underlined}.}
\label{tab:nvidia}
\setlength{\tabcolsep}{0pt} 
\begin{tabular*}{\linewidth}{@{\extracolsep{\fill}} l cc p{0.2em} l cc} 
\toprule
Method & PSNR $\uparrow$ & LPIPS $\downarrow$ & \phantom{abc} & Method & PSNR $\uparrow$ & LPIPS $\downarrow$ \\
\cmidrule{1-3} \cmidrule{5-7} 
D-NeRF~\cite{pumarola2021d}    & 21.49 & 0.232 & & CTNeRF~\cite{miao2024ctnerf}          & 26.13 & 0.082 \\
NR-NeRF~\cite{tretschk2021non} & 19.69 & 0.323 & & DynPoint~\cite{zhou2024dynpoint}      & 26.53 & \underline{0.068} \\
TiNeuVox~\cite{fang2022fast}   & 19.74 & 0.285 & & D-NPC~\cite{kappel2024d}              & 25.64 & 0.109 \\
HyperNeRF~\cite{park2021hypernerf} & 17.60 & 0.367 & & RoDynRF~\cite{rodynrf}             & 25.89 & \textbf{0.067} \\
NSFF~\cite{li2021neural}       & 24.33 & 0.199 & & Casual-FVS~\cite{lee2023fast}         & 24.57 & 0.081 \\
DynNeRF~\cite{gao2021dynamic}  & 26.10 & 0.082 & & GaussianMarbles~\cite{stearns2024dynamic} & 22.32 & 0.129 \\
MonoNeRF~\cite{tian2023mononerf} & 25.62 & 0.106 & & MoSca~\cite{lei2024mosca}            & \underline{26.72} & 0.070 \\
4DGS~\cite{wu20244d}           & 21.45 & 0.199 & & \cellcolor[gray]{.9}\ourmodel{} (Ours)           & \cellcolor[gray]{.9}\textbf{26.75} &\cellcolor[gray]{.9} \underline{0.068} \\

\bottomrule
\end{tabular*}
\vspace{-1em} 
\end{table}

\begin{figure*}[tbp] 
\centering
\includegraphics[width=\textwidth]{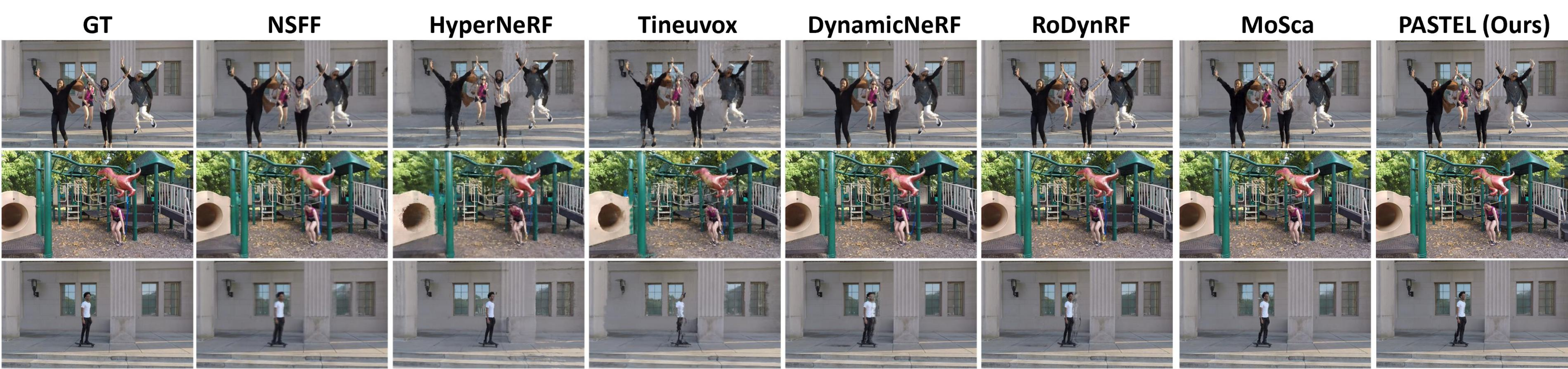} 
\vspace{\figvspace}
\caption{\textbf{Qualitative comparison on NSFF~\cite{yoon2020novel}.} Our method achieves better overall quality and consistency with the ground truth. Please zoom in for clearer visualization.}
\label{fig:nvidia} 
\vspace{\figvspace}
\end{figure*}

\begin{figure*}[ht] 
\centering
\includegraphics[width=\textwidth]{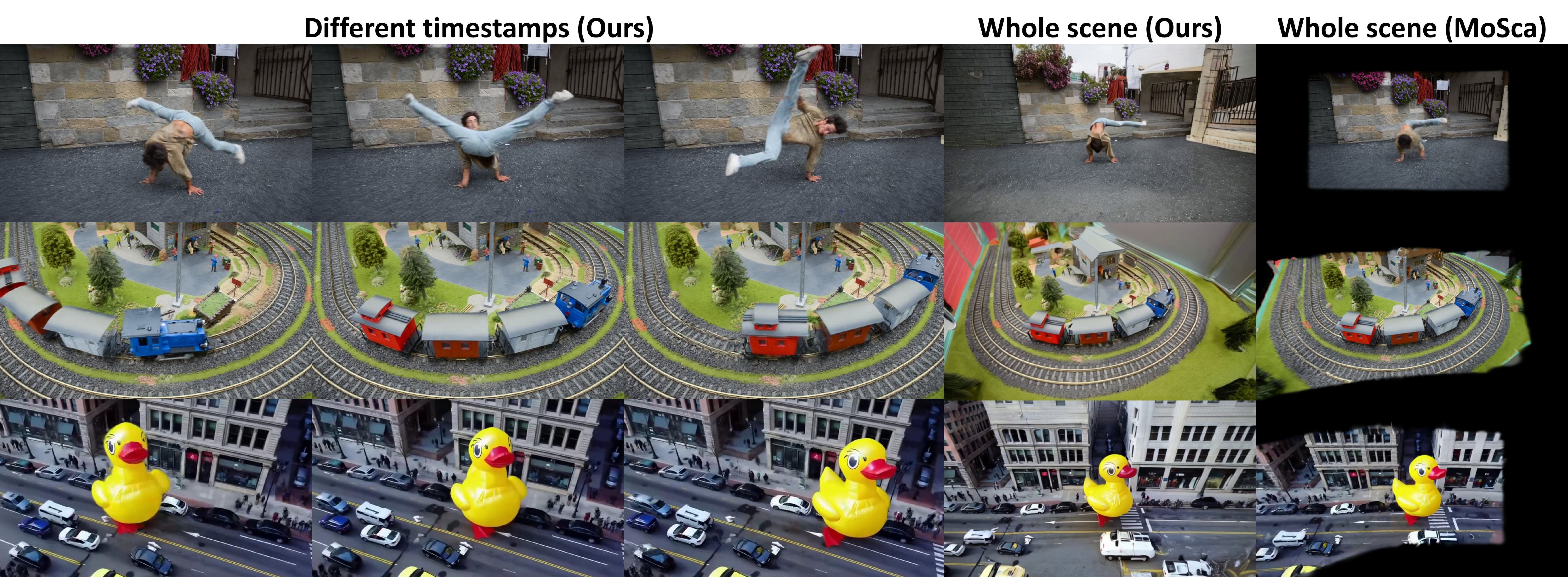} 
\caption{\textbf{Reconstruction on real-world monocular videos.} The first three columns exhibit novel view synthesis with different timestamps. The last two columns compare the whole scene with the previous state-of-the-art method, Mosca~\cite{lei2024mosca}. \ourmodel{} can be effectively applied to the reconstruction of any monocular video, achieving both superior overall quality and effective synthesis of extended views.}
\label{fig:random_videos} 
\end{figure*}

We further compare \ourmodel{} against a stronger Cartesian baseline (Cartesian-TopK) that selects coverage-maximizing views from 100 random trajectories. As shown in \cref{fig:better_baseline}, our panoramic trajectory planning still yields more structured and consistent view expansions.
\begin{figure}[t]
\centering
\includegraphics[width=\linewidth]{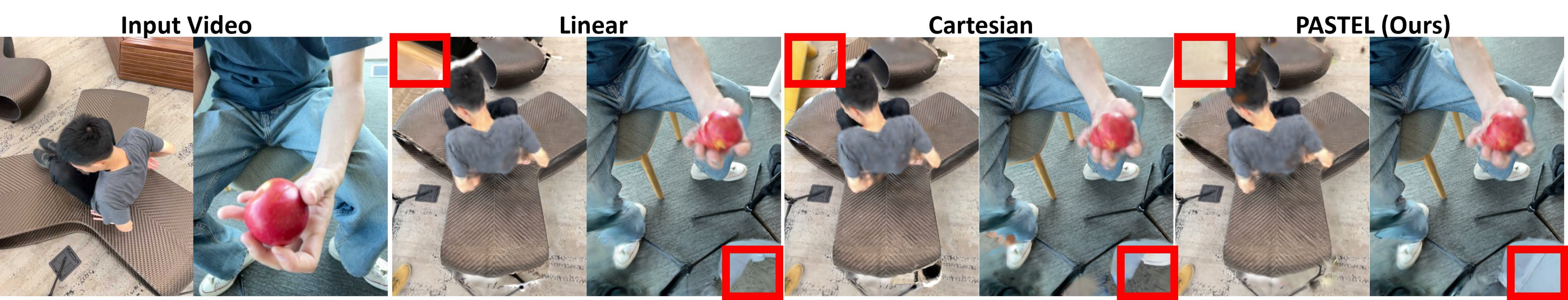}
\caption{\textbf{Qualitative comparison with Cartesian-TopK baseline.} Our panoramic trajectory planning provides more structured and consistent view expansions than unconstrained Cartesian sampling.}
\label{fig:better_baseline}
\end{figure}

\section{More results on real-world monocular videos}
\label{sec:real-world}

Dynamic evaluation focuses on objects undergoing complex motion, self-occlusion, or disocclusion. Current datasets such as DyCheck\cite{li2023dynibar} and NSFF\cite{yoon2020novel} rarely contain dynamic content that moves beyond observable boundaries. Moreover, highly dynamic regions fully unobservable cannot be reconstructed reliably by any method. Their motions are completely unobservable from input monocular video.  Therefore, we focus our analysis on partially occluded dynamic regions. 

To evaluate on the adaptation of \ourmodel{} to other complex real-world monocular videos, we experiment on casual monocular videos of dynamic scenes with irregular motion and crowded environments. As illustrated in \cref{fig:random_videos}, \ourmodel{} enables adaptation to irregular movements of the complex hip hop in the first row, heavy occlusion cases of the train in the second row, and crowded environments in the third row. Besides, it is evident that previous reconstruction-based methods\cite{lei2024mosca} frequently encounter limitations when attempting to extend beyond observable boundaries. In stark contrast, \ourmodel{} successfully facilitates the generation of novel views while maintaining high fidelity in novel-view synthesis.

\section{More ablations}
\label{sec:ablation}
We conduct more ablations of our method on the IPhone dataset\cite{li2023dynibar}.

\subsection{Hyperparameter sensitivity.}
We further analyze sensitivity to trajectory directions $M$, interpolation steps $L$, and SSIM threshold $\epsilon$, as shown in Table~\ref{tab:hyperparameters} and \cref{fig:hyperparameters}.
Small $M$ yields insufficient scene exploration. An excessively large $M$ causes multiple novel trajectories to collide, degrading consistency.
For interpolation steps $L$ in Eq. 7, too few steps create overly abrupt viewpoint transitions, whereas excessively large $L$ requires multiple rounds of diffusion refinement and introduces inconsistency.
Finally, the SSIM threshold $\epsilon$ in Eq. 10 regulates the ratio of reliable to unreliable regions.
A small $\epsilon$ suppresses the refinement of generative priors, while a large $\epsilon$ causes unnecessary supervision from synthesized frames.

\begin{table}[t]
\centering
\caption{\textbf{Ablations on trajectory directions $M$, interpolation steps $L$, and SSIM threshold $\epsilon$.} 
Gray rows indicate the default settings used in our final model.}
\label{tab:hyperparameters}
\resizebox{\linewidth}{!}{
\begin{tabular}{cccc c cccc c cccc}
\toprule
\multicolumn{4}{c}{Trajectory Directions ($M$)} & \phantom{a} & 
\multicolumn{4}{c}{Interpolation Steps ($L$)} & \phantom{a} & 
\multicolumn{4}{c}{SSIM Threshold ($\epsilon$)} \\
\cmidrule{1-4} \cmidrule{6-9} \cmidrule{11-14}
$M$ & PSNR & SSIM & LPIPS && $L$ & PSNR & SSIM & LPIPS && $\epsilon$ & PSNR & SSIM & LPIPS \\
\midrule
2 & 17.70 & 0.573 & 0.355 && 12 & 17.88 & 0.575 & 0.352 && 0.1 & 18.29 & 0.569 & 0.350 \\
\rowcolor[gray]{.9}
8 & \textbf{18.54} & \textbf{0.582} & \textbf{0.335} && 49 & \textbf{18.54} & \textbf{0.582} & \textbf{0.335} && 0.3 & \textbf{18.54} & \textbf{0.582} & \textbf{0.335} \\
32 & 17.98 & 0.578 & 0.342 && 196 & 17.61 & 0.573 & 0.357 && 0.5 & 18.06 & 0.560 & 0.358 \\
\bottomrule
\end{tabular}
}
\end{table}
\begin{figure*}[ht] 
\centering
\includegraphics[width=\textwidth]{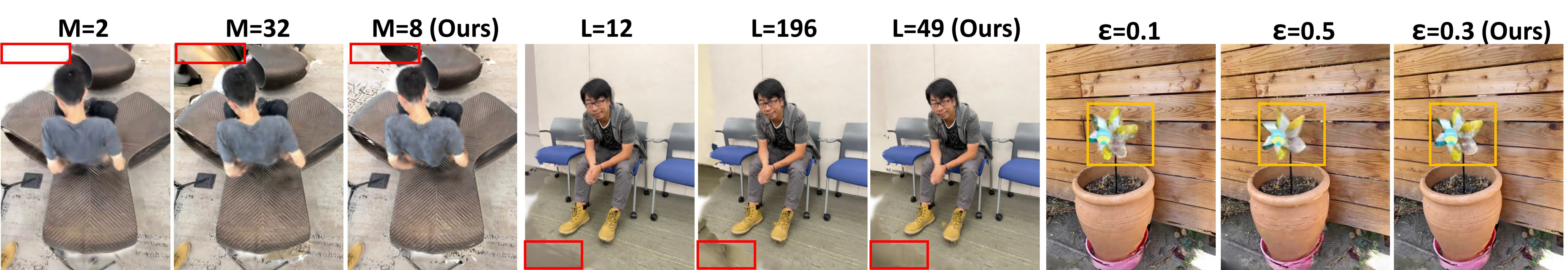}
\vspace{\figvspace}
\caption{\textbf{ Ablations on trajectory directions $M$, interpolation steps $L$, and SSIM threshold $\epsilon$.} Trajectory directions $M$ and interpolation steps $L$ has significant influence on red-boxed regions beyond observable boundary. SSIM threshold $\epsilon$ influences detailed textures on orange-boxed regions.
}
\label{fig:hyperparameters} 
\vspace{\figvspace}
\end{figure*}

\begin{table}[t]
\centering
\caption{\textbf{Ablation study on generative priors.} We compared three recently open-sourced camera-controlled video generation methods. TrajectoryCrafter performs best in our architecture.}
\label{tab:abl_prior}
\begin{tabular}{l ccc}
\toprule
Generative prior & PSNR$\uparrow$ & SSIM$\uparrow$ & LPIPS$\downarrow$ \\
\midrule
\rowcolor[gray]{.95} 
TrajectoryCrafter~\cite{yu2025trajectorycrafter} & \textbf{18.54} & \textbf{0.582} & \textbf{0.335} \\
ReCamMaster~\cite{bai2025recammaster}            & 17.95 & 0.551 & 0.361 \\
ViewCrafter~\cite{yu2024viewcrafter}             & 17.22 & 0.513 & 0.408 \\
\bottomrule
\end{tabular}
\end{table}

\begin{table}[t]
\centering
\caption{\textbf{Ablation study on robustness to estimation errors.} 
We simulate inaccuracy by adding noise with standard deviation (std) to the original depth (left) and camera pose (right).}
\label{tab:robustness}
\resizebox{\linewidth}{!}{
\begin{tabular}{c c c}
\begin{minipage}{0.48\linewidth}
    \centering
    \textbf{(a) Inaccurate Depth} \\
    \begin{tabular}{l ccc}
    \toprule
    Depth std & PSNR$\uparrow$ & SSIM$\uparrow$ & LPIPS$\downarrow$ \\
    \midrule
    \cellcolor[gray]{.95} 0\% & \cellcolor[gray]{.95}\textbf{18.54} & \cellcolor[gray]{.95}\textbf{0.582} & \cellcolor[gray]{.95}\textbf{0.335} \\
    1\%  & 18.29 & 0.569 & 0.350 \\
    5\%  & 17.40 & 0.532 & 0.394 \\
    20\% & 15.24 & 0.426 & 0.500 \\
    \bottomrule
    \label{tab:abl_depth}
    \end{tabular}
\end{minipage}
&
\phantom{sp}
&
\begin{minipage}{0.48\linewidth}
    \centering
    \textbf{(b) Inaccurate Pose} \\
    \begin{tabular}{l ccc}
    \toprule
    Pose std & PSNR$\uparrow$ & SSIM$\uparrow$ & LPIPS$\downarrow$ \\
    \midrule
    \cellcolor[gray]{.95} $0^\circ$ & \cellcolor[gray]{.95}\textbf{18.54} & \cellcolor[gray]{.95}\textbf{0.582} & \cellcolor[gray]{.95}\textbf{0.335} \\
    $1^\circ$    & 18.26 & 0.567 & 0.346 \\
    $5^\circ$    & 17.17 & 0.505 & 0.412 \\
    $20^\circ$   & 14.63 & 0.395 & 0.545 \\
    \bottomrule
        \label{tab:abl_pose}
    \end{tabular}
\end{minipage}
\end{tabular}
}
\end{table}

\subsection{Ablation on generative priors} 
\label{sec:abl_gen_prior}
Table~\ref{tab:abl_prior} shows performance variations when using different camera-controlled video generation models as generative priors. Among them, TrajectoryCrafter\cite{yu2025trajectorycrafter} proves to be the most effective, likely due to its precise control derived from depth and pose projections. ReCamMaster\cite{bai2025recammaster} shows slightly lower performance, possibly because of its deviation in pose estimation. ViewCrafter\cite{yu2024viewcrafter} exhibits lower PSNR, since it cannot support dynamic generation. Nevertheless, TrajectoryCrafter\cite{yu2025trajectorycrafter} still has certain inconsistencies, as shown in Fig. 3 of the main paper.  Improving the consistency of generative priors is still crucial for further enhancing our model's performance.


\subsection{Sensitivity analysis on depth and pose perturbations} 
\label{sec:abl_depth}
As shown in Table~\ref{tab:abl_depth} and Table~\ref{tab:abl_pose}, \ourmodel{} maintains strong performance under slight perturbations in depth ($1\%$) and pose ($1^\circ$ ) estimation. However, when the estimation errors become significant ($20\%$ for depth and $20^\circ$ for pose), the performance drops noticeably. It illustrates the reliance of \ourmodel{} on accurate depth and pose estimations.

\section{Scene-specific adaptation}
To evaluate on scene-specific adaptation, we fine-tune our diffusion prior using only observed regions of the input monocular video. 
Specifically, we exploit TrajectoryCrafter\cite{yu2025trajectorycrafter} as generative prior. We finetune the cross-attention and patch embedding layers in the Ref-DiT blocks of our generative prior while freezing all other parameters on monocular input videos. 

As shown in \cref{fig:finetuning}, 
fine-tuning yielded a slight improvement in color consistency for extrapolated views. We hypothesize this is because these unexplored regions remain entirely unobserved in the monocular input. Therefore, scene-specific adaptation cannot provide reliable geometric supervision for them.
Moreover, \ourmodel{} already conditions its generative prior on monocular observations. The generative prior have effectively conditioned on the scene without further fine-tuning. Therefore, additional scene-specific fine-tuning thus brings limited improvement.

\begin{figure*}[t] 
\centering
\includegraphics[width=\textwidth]{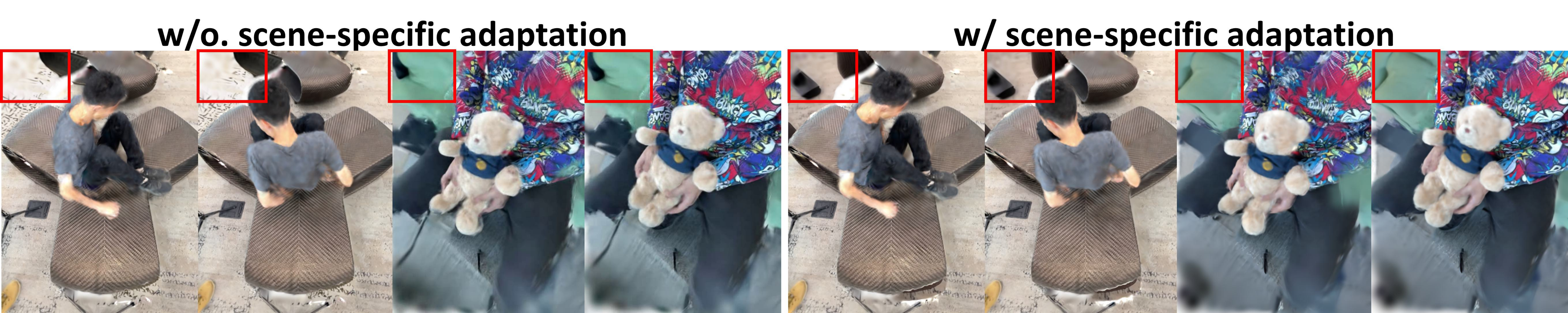} 
\caption{\textbf{ Scene-specific adaptation of the generative prior.} Fine-tuning on observed regions shows only slight improvement of color consistency in red-boxed unseen regions.}
\label{fig:finetuning} 
\end{figure*}

\section{Limitations.}
\label{sec:limitation}
Despite the high fidelity and consistency of our reconstruction results, our method still relies on accurate depth, pose, and flow estimations, as well as stable camera-controlled video generation priors. 
Furthermore, the core geometrical assumption of our panoramic representation has limitations under extremely large camera motions. 
Handling arbitrary viewing angles also remains highly challenging. Additionally, there remains a domain gap between the true reconstruction and the generative output. Generated regions beyond the visible boundaries occasionally appear noticeably blurry and lack high-frequency texture details compared to the original covisible regions.
On the other hand, \cite{zhang2024monst3r,wang2025vggt,wang2025pi} recently proposed feed-forward neural networks that directly infer all key 3D attributes of a scene, including depth and pose estimation. However,
their requirement for GPU memory is directly proportional to the frame number of the monocular video. Consequently, it is challenging to apply these methods directly to monocular videos.  
Meanwhile, some concurrent works\cite{wu2025video,yesiltepe2025dynamic} recently achieved improved performance on camera-controlled video generation. We also plan to explore these works as generative priors after their code release.
Another promising research direction would be to design a feed-forward network for directly reconstructing both observable boundaries and unbounded content beyond observable boundaries. This feed-forward network may be achieved by delicate integration of both feed-forward neural networks\cite{zhang2024monst3r,wang2025vggt,wang2025pi} and video generation models\cite{yu2025trajectorycrafter, bai2025recammaster,wu2025video,yesiltepe2025dynamic}.

\fi

\end{document}